\documentclass[runningheads]{llncs}
\usepackage[T1]{fontenc}
\usepackage{graphicx}
\usepackage{amsmath}
\usepackage{xcolor}
\usepackage{amsmath}
\usepackage{amssymb}
\usepackage{float}
\usepackage{multirow}
\usepackage{caption} 
\usepackage{dblfloatfix}
\usepackage{array}
\usepackage{marvosym}
\makeatother

\begin{document}

\title{A Self-Supervised Image Registration Approach for Measuring Local Response Patterns in Metastatic Ovarian Cancer}

\titlerunning{A Self-Supervised Registration Approach for Metastatic Ovarian Cancer}


\author{Inês P. Machado*\inst{1,2,3}\textsuperscript{(\Letter)} \and
Anna Reithmeir\inst{4,5}  \and
Fryderyk Kogl\inst{4,5}  \and
Leonardo Rundo\inst{6} \and
Gabriel Funingana\inst{1,2,3} \and
Marika Reinius\inst{1,2,3} \and
Gift Mungmeeprued\inst{1,2,3} \and
Zeyu Gao\inst{1,2,3} \and
Cathal McCague\inst{1,2} \and
Eric Kerfoot\inst{7} \and
Ramona Woitek\inst{8} \and
Evis Sala\inst{9} \and
Yangming Ou\inst{10} \and
James Brenton\inst{1,2} \and
Julia Schnabel\inst{4,5,7}\and
Mireia Crispin\inst{1,2,3}}

\authorrunning{I.P. Machado et al.}

\institute{Department of Oncology, University of Cambridge, UK \and 
Cancer Research UK, Cambridge Institute, University of Cambridge, UK\and
Early Cancer Institute, University of Cambridge, UK \and
School of Computation, Information \& Technology, Technical University of Munich \and
Institute of Machine Learning in Biomedical Imaging, Helmholtz Munich, Germany \and
Department of Information and Electrical Engineering, University of Salerno, Italy \and
School of Biomedical Engineering \& Imaging Sciences, King’s College London, UK \and
Research Center for Medical Image Analysis and AI, Danube University, Austria \and
Department of Radiologic Sciences, Università Cattolica del Sacro Cuore, Italy \and
Department of Radiology, Boston Children's Hospital, Harvard Medical School \\
\email{\{im549\}@cam.ac.uk}}

\maketitle

\vspace{-2em}
\begin{abstract} 
High-grade serous ovarian carcinoma (HGSOC) is characterised by significant spatial and temporal heterogeneity, typically manifesting at an advanced metastatic stage. A major challenge in treating advanced HGSOC is effectively monitoring localised change in tumour burden across multiple sites during neoadjuvant chemotherapy (NACT) and predicting long-term pathological response and overall patient survival. In this work, we propose a self-supervised deformable image registration algorithm that utilises a general-purpose image encoder for image feature extraction to co-register contrast-enhanced computerised tomography scan images acquired before and after neoadjuvant chemotherapy. This approach addresses challenges posed by highly complex tumour deformations and longitudinal lesion matching during treatment. Localised tumour changes are calculated using the Jacobian determinant maps of the registration deformation at multiple disease sites and their macroscopic areas, including hypo-dense (i.e., cystic/necrotic), hyper-dense (i.e., calcified), and intermediate density (i.e., soft tissue) portions. A series of experiments is conducted to understand the role of a general-purpose image encoder and its application in quantifying change in tumour burden during neoadjuvant chemotherapy in HGSOC. This work is the first to demonstrate the feasibility of a self-supervised image registration approach in quantifying NACT-induced localised tumour changes across the whole disease burden of patients with complex multi-site HGSOC, which could be used as a potential marker for ovarian cancer patient’s long-term pathological response and survival. 
\vspace{-0.6em}
\end{abstract}

\section{Introduction}

Despite notable advances in the treatment of ovarian cancer, it continues to be one of the leading causes of gynaecological cancer death among women~\cite{HGSOC1}. High-grade serous ovarian carcinoma (HGSOC) is the most common type and is a highly heterogeneous disease that typically presents with advanced, multi-site metastatic disease. Neoadjuvant chemotherapy (NACT) followed by delayed primary surgery is becoming the most frequent treatment strategy for advanced HGSOC~\cite{HGSOC}. HGSOC often exhibits lesion-wise heterogeneous response to NACT with some lesions disappearing or shrinking, some remaining stable, some growing, and new lesions appearing despite ongoing treatment. The response of individual lesions to treatment has been shown to drive progression and provides predictive information for the effectiveness of delayed primary surgery and survival~\cite{ICON8_clinical_trial}. Consequently, treatment response assessment following NACT is critical to properly plan subsequent stages of the patient's clinical pathway. 

In current clinical practice, tumour response assessment is usually based on two-dimensional measurements to approximate change in tumour size on serial contrast-enhanced computerised tomography (CE-CT) scan images. Although Response Evaluation Criteria in Solid Tumours (RECIST 1.1) is the current clinical guideline to assess size change of solid tumours after therapeutic treatment, it is based on measurements of lesion diameters, and therefore is not able to capture the three-dimensional complexity of tumours. Additionally, radiologists have to choose which target lesions to track and measure, which adds subjectivity to the clinical evaluation process. 

\begin{figure}[t]
\includegraphics[trim={0 5cm 9.6cm 3cm},width=\textwidth]{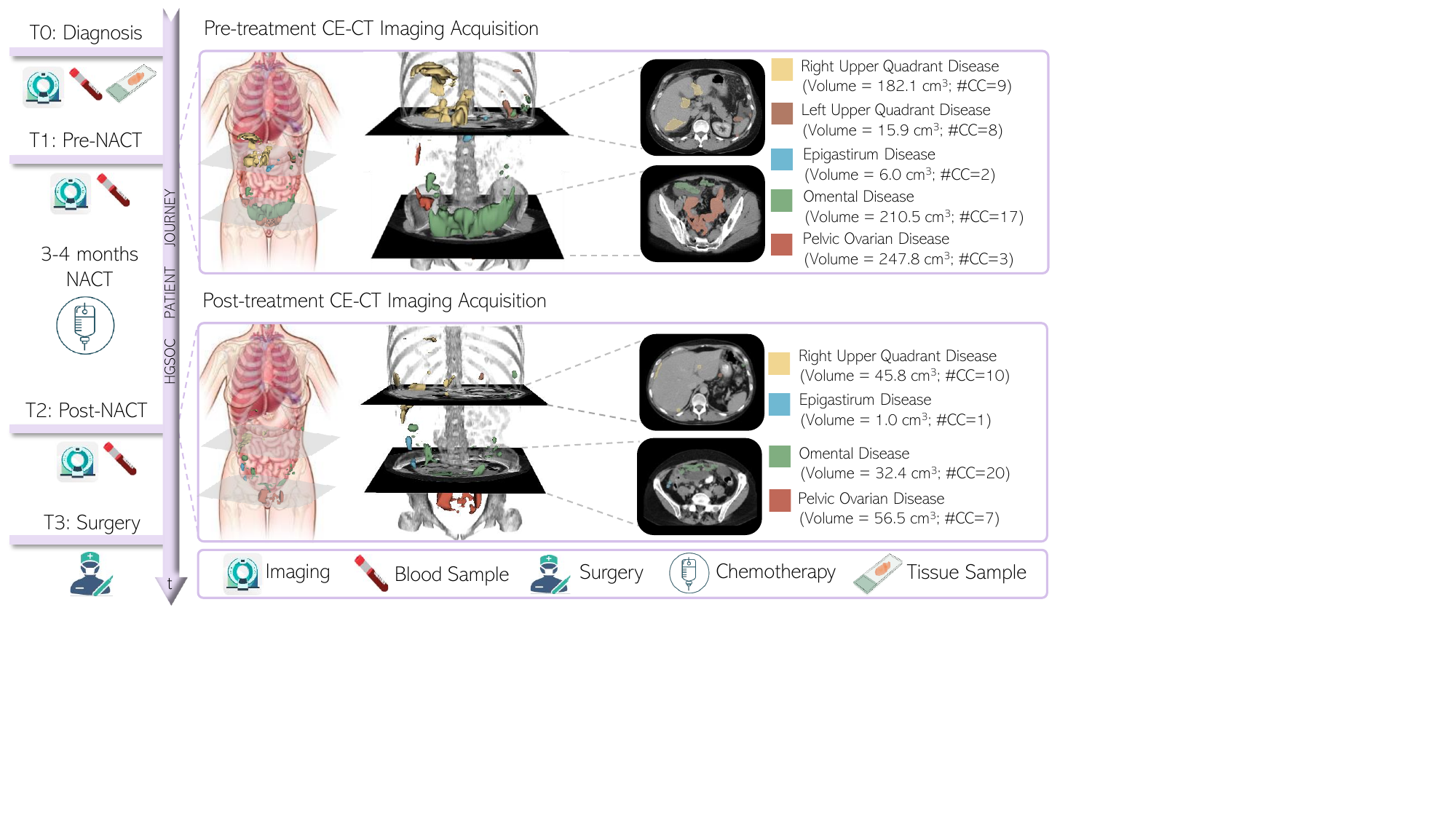}
\caption{Overview of the treatment schedule and characteristics of an advanced-stage high-grade serous carcinoma. Initial staging and diagnosis were based on CE-CT imaging and biopsy. Surgery was performed after two cycles of NACT. According to RECIST 1.1 criteria, the patient was classified as a partial responder. Pre- and post-treatment CE-CT scans revealed the metastatic nature of the disease, identifying 16 sites of disease burden with varying numbers of connected components (CC) and highlighting heterogeneity in lesion response. } \label{disease}
\end{figure}

Future clinical trials for ovarian cancer require stronger, more sensitive response metrics that can provide more robust discrimination between different response patterns. Here, a first challenge is how to effectively monitor volumetric tumour changes across multiple disease sites during the course of NACT and, more importantly, how to effectively reveal heterogeneous changes in various sub-regions within tumours rather than tumours as a whole, since the heterogeneity may be more indicative of long-term pathological response. A second challenge is how to match multiple lesions between longitudinal scans. Lesions may be numerous and densely packed within a single organ or tissue. For example, metastatic ovarian cancer patients with a high disease burden can exhibit dozens of disease sites with hundreds of small connected components~\cite{Mireia}. Additionally, lesions may grow, shrink, split, or merge over time. Thus, matching lesions between longitudinal images is critical to ensure that heterogeneous response patterns can be identified, and imaging data can be best utilised to inform treatment decisions. Figure~\ref{disease} illustrates pre- and post-treatment CE-CT scans that highlight the metastatic nature of HGSOC and the lesion-wise heterogeneity in response to chemotherapy, with some lesions disappearing or shrinking, some remaining stable, some growing, and new lesions appearing.

Registration of longitudinal CE-CT images acquired during the course of treatment has received increasing interest as a useful tool to enable localised lesion tracking during therapy. Despite its importance, longitudinal registration of ovarian cancer CE-CT scans has been less studied, and the registration performance is less understood compared to other organs (e.g., lung \cite{Lung1,Lung2}, liver \cite{Liver1,Liver2}, brain \cite{Brain1}, prostate \cite{Prostate1}, cervix \cite{Cervical1} and head and neck \cite{HeadNeck1}).

With the emergence of deep learning, there has been a notable shift in image registration techniques. These approaches directly predict a displacement field by leveraging self-learned hierarchical features extracted from both the input moving and reference images. In particular, the introduction of foundation models in computer vision marks a significant paradigm shift in the approach to complex visual tasks. Self-supervised vision foundation models demonstrate this evolution by providing pre-trained models that have learned rich feature representations from large-scale unlabeled datasets. This is particularly advantageous in scenarios like HGSOC, where little data is available and no publicly available imaging datasets exist. In such cases, self-supervised learning becomes instrumental in overcoming data scarcity challenges. \newline
\indent This work proposes a deformable image registration framework based on a general-purpose image encoder to assess tumour changes over time as a measure of treatment response in metastatic HGSOC. The assessment of registration performance covers various anatomical structures in the pelvic, abdominal, and chest regions. Specifically, it examines registration accuracy among disease sites and patient subgroups with different treatment responses. Volumetric disease change is quantified for the first time for each lesion individually, and for different sub-regions within tumours. Our ultimate goal is to use registration-quantified tumour changes in HGSOC to more accurately measure treatment response and predict long-term pathological response and patient survival after treatment.

\vspace{-10pt}
\section{Methods}
\begin{figure}[t]
\includegraphics[trim={0 1cm 4cm 1cm},width=\textwidth]{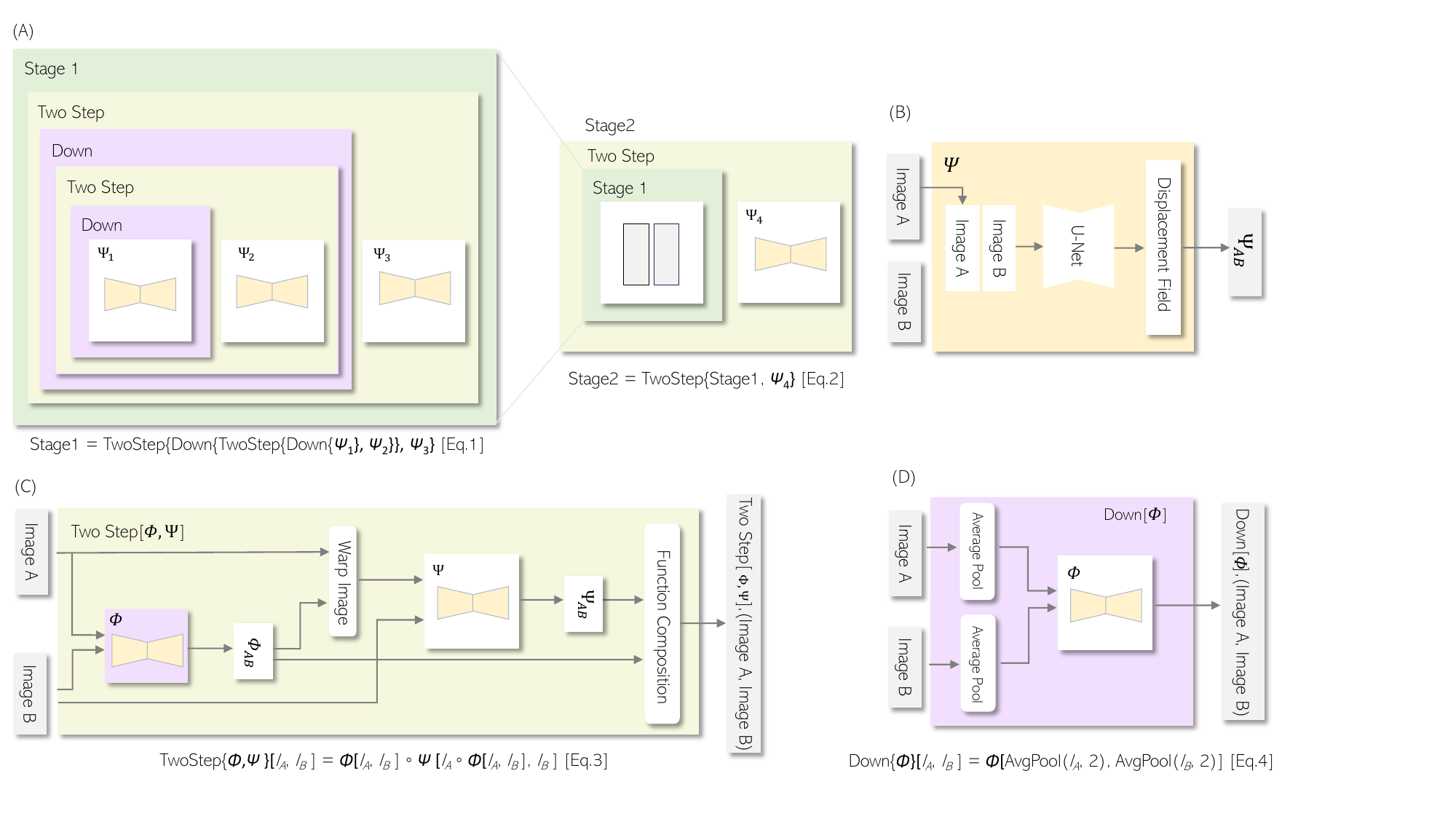}
\caption{Image registration follows a multi-step, multi-resolution approach trained with a two-stage process (A) via the Two Step (C) and the Downsample (D) operators. The basic component $\Psi$ of the registration network is represented by a U-Net instance, taking as input two images and returning a displacement field (B). The downsample operator predicts the warp between two high-resolution images using a network that operates on low-resolution images, and the two-step operator predicts the warp between two images in two steps, first capturing the coarse transform via $\Phi$ and then the residual transform via $\Psi$.} \label{pipeline}
\end{figure}

\vspace{-3pt}
\noindent\textbf{Registration framework.} The source and target images in the registration problem are denoted by $I_A : \Omega \rightarrow \mathbb{R}$ and $I_B : \Omega \rightarrow \mathbb{R}$, respectively.  The transformation map is denoted by $\Phi_{AB} : \mathbb{R}^d \rightarrow \mathbb{R}^d$ with the intention that $I_A \circ \Phi_{AB} \sim I_B$. We adopt the GradICON~\cite{GradICON} registration network which uses a two-step registration process. In the first step, images are passed through a three-level multiresolution registration network. At each level, a U-Net accepts the warped image from the previous level and the target image. Additionally, the input images are downsampled to \(\frac{1}{4}\) and \(\frac{1}{2}\) for each level, respectively. In the second step, images are passed through a single U-Net, which takes the warped image from the first step and the target image at full resolution. All four U-Nets share the same architecture as illustrated in Figure~\ref{pipeline}. \\

\vspace{-5pt}
\noindent\textbf{Volumetric feature encoder and training.} We adopt the publicly-available uniGradICON~\cite{uniGradICON} feature encoder, a foundation model trained and evaluated on 12 public medical image datasets from the Learn2Reg Grand Challenge\footnote[1]{ https://learn2reg.grand-challenge.org/}, covering various anatomical regions, imaging modalities, and deformation patterns. Given an ordered image pair $(I_A, I_B)$, the registration network outputs the transformation map $\Phi_{AB}$ which maps $I_A$ to the space of $I_B$. By swapping the input pair $(I_B, I_A)$, we obtain the estimated inverse map $\Phi_{BA}$. The similarity loss $L_{\text{sim}}$ is computed between the warped image $I_A \circ \Phi_{AB}$ and the target image $I_B$, and vice versa following the formulation:
\vspace{-3pt}
\begin{equation*}
L = L_{\text{sim}}(I_A \circ \Phi_{AB}, I_B) + L_{\text{sim}}(I_B \circ \Phi_{BA}, I_A) + \lambda \left\| \nabla (\Phi_{AB} \circ \Phi_{BA}) - I \right\|^2_F
\end{equation*}

\vspace{-3pt}
\noindent We use localised normalised cross correlation (1 - LNCC) as the similarity measure following the implementation described in~\cite{GradICON}. The third term is the gradient inverse consistency regularizer, which penalizes differences between the Jacobian of the composition of $\Phi_{AB}$ with $\Phi_{BA}$ and the identity matrix $I$; $\|\cdot\|_F^2$ is the Frobenius norm, $\lambda > 0$. The first and second stages of the registration network were trained for 800 and 200 epochs, respectively, with a learning rate of \(5 \times 10^{-5}\). \\

\vspace{-5pt}
\noindent\textbf{Quantification of volumetric tumour progression.} The obtained deformation vector fields were further used to calculate the Jacobian maps. 
The Jacobian map was computed as the determinant of the gradient of the deformation vector field (DVF). The Jacobian determinants indicate the volumetric change ratio at each voxel, where a value greater than 1 indicates voxel expansion, a value less than 1 indicates shrinkage, and a value of exactly 1 indicates volume preservation. 
\vspace{2mm}




\noindent\textbf{Intra- and inter-tumoural heterogeneity.} HGSOC is characterised by high levels of macroscopic heterogeneity, with frequent hyper-dense (calcified), hypo-dense (cystic/necrotic), or intermediate dense (soft tissue) tumour regions, each of which can have different prognostic significance and influence therapy response. To understand how different tumour regions respond to treatment, an automated sub-segmentation of these regions was performed using a tissue-specific sub-segmentation framework based on an unsupervised fuzzy clustering technique previously described in~\cite{Sub_segmentations}. Jacobian maps were computed as the determinant of the gradient of the deformation vector field to quantify local changes for each tissue-specific sub-segmentation. 

\vspace{-12pt}
\section{Experiments}
\begin{figure}[!]
\includegraphics[width=\textwidth]{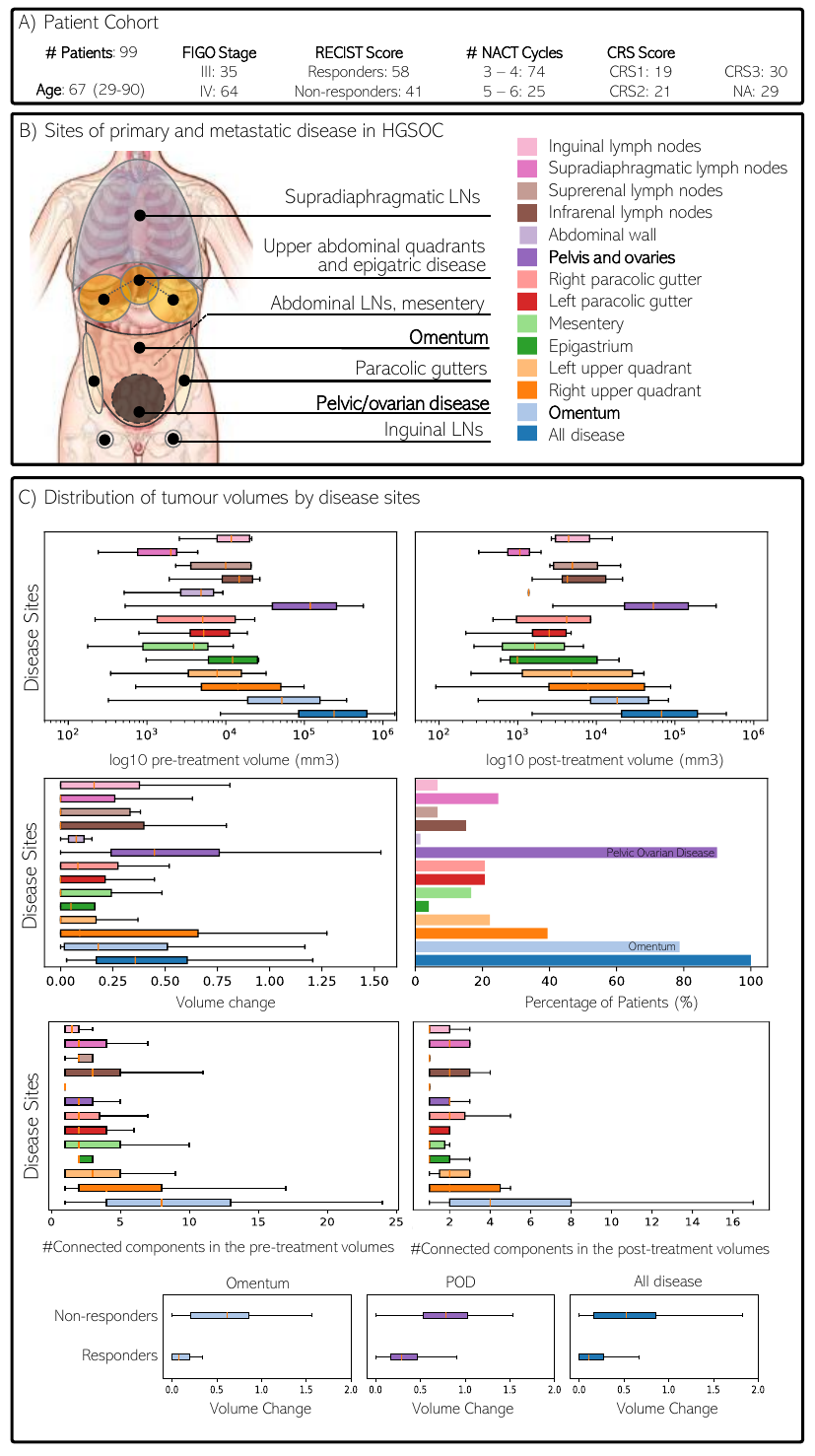}
\caption{ (A) Main characteristics of the patient cohort. (B) Sites of primary and metastatic HGSOC disease. (C) Distribution of tumour volumes by disease site.}\label{data}
\end{figure}

\vspace{-5pt}
\textbf{Patient cohort.} A total of 99 patients with a confirmed histopathological diagnosis of HGSOC, who underwent neoadjuvant chemotherapy before delayed primary surgery, were included in this study. All patients were treated at Cambridge University Hospitals NHS Foundation Trust and were recruited into a clinical observational study approved by the local research ethics committee (REC reference number: 08/H0306/61). Clinical CE-CT scans covering the abdomen and pelvis were acquired. Figure~\ref{data} (a) shows the median (min - max) age of the patient cohort in years, the total number of NACT cycles administered, and the FIGO stage (I to IV), which describes how far the cancer has spread. The radiological response is measured based on the RECIST 1.1 criteria and is categorised into four groups: complete response (CR), partial response (PR), stable disease (SD), or progressive disease (PD). Chemotherapy Response Scores (CRS) describe the histopathologic response to NACT in the omentum (fatty tissue layer in the abdominal cavity) and are categorized into I to III.

\noindent\textbf{Image segmentation and labelling.} On CE-CT axial images reconstructed with a slice thickness of typically 5 mm, all cancer lesions were segmented semi-automatically by a board-certified radiologist with ten years of experience in clinical imaging, using Microsoft Radiomics (project InnerEye; Microsoft, Redmond, WA, USA). The volumes of interest were annotated for their anatomic location with the following disease sites and volume percentages: ovaries and pelvis (66.4\%), omentum (24.4\%), right upper quadrant (5.1\%), left upper quadrant (1.1\%), epigastrium (0.1\%), mesentery (0.3\%), right paracolic gutter (0.8\%), left paracolic gutter (0.3\%), infrarenal abdominal lymph nodes (0.8\%), suprarenal abdominal lymph nodes (0.2\%), inguinal lymph nodes (0.2\%), and supradiaphragmatic lymph nodes (0.1\%). Cystic and solid tumour parts were included in these segmentations. Intra-tumoural sub-regions were automatically segmented as described in~\cite{Sub_segmentations}. Tumour in the omentum and pelvic/ovarian locations accounts for the majority of the disease burden at baseline and are the most frequent tumour locations. Figure~\ref{data} (b) and (c) show the sites of primary and metastatic disease in HGSOC and the distribution of tumour volumes, volume change (follow-up/ baseline) and number of connected components by disease site, as well as the total, omental and pelvic/ovarian disease volume change stratified by RECIST 1.1 response status. Additionally, segmentations of 10 anatomical structures, including left and right kidneys, liver, spleen, lungs, stomach, pancreas, colon, vertebra and sacrum in the CE-CT images were automatically computed for each scan using the publicly-available segmentation tool Total Segmentator v2~\cite{TotalSegmentator} and used for registration evaluation. Baseline and follow-up CE-CT scans were evaluated according to RECIST 1.1 for treatment response assessment. 

\noindent\textbf{Comparative approaches and evaluation criteria.} The proposed foundation model was evaluated against two state-of-the-art techniques: NiftyReg~\cite{NiftyReg}, a conventional registration technique, and VoxelMorph~\cite{VoxelMorph}, the latest benchmark for deformable image registration. To provide a comprehensive evaluation of the registration performance, we consider a number of complementary metrics that assess the accuracy, plausibility, and speed of the proposed approach. The agreement between registration-propagated contours and ground-truth contours was assessed with the Dice Similarity Coefficient (DSC) for tumour and non-tumour regions. The spatial Jacobian and deformation vector field images were used to assess the plausibility of the resulting transformations. The plausibility (smoothness) of a displacement field is captured using the standard deviation of the logarithm of the Jacobian determinant (SDlogJ) of the displacement field~\cite{SDlogJ}. Visual inspection of the deformed images was also performed to detect any large, noticeable issues with the transformation. In addition, we measure the test-time registration runtime (RT). The registration method obtaining the highest accuracy was used to quantify localised volumetric tumour changes in HGSOC patients undergoing NACT. 

\section{Results}

\noindent\textbf{Registration performance.} Table 1 (A) and (B) show the Dice coefficient and the standard deviation of the logarithm of the Jacobian determinant of different registration methods across different disease sites (A), anatomical structures (B), and patient subgroups with different treatment responses. \\

\begin{figure}[!]
\includegraphics[trim={0 4.5cm 5.3cm 3cm},width=\textwidth]{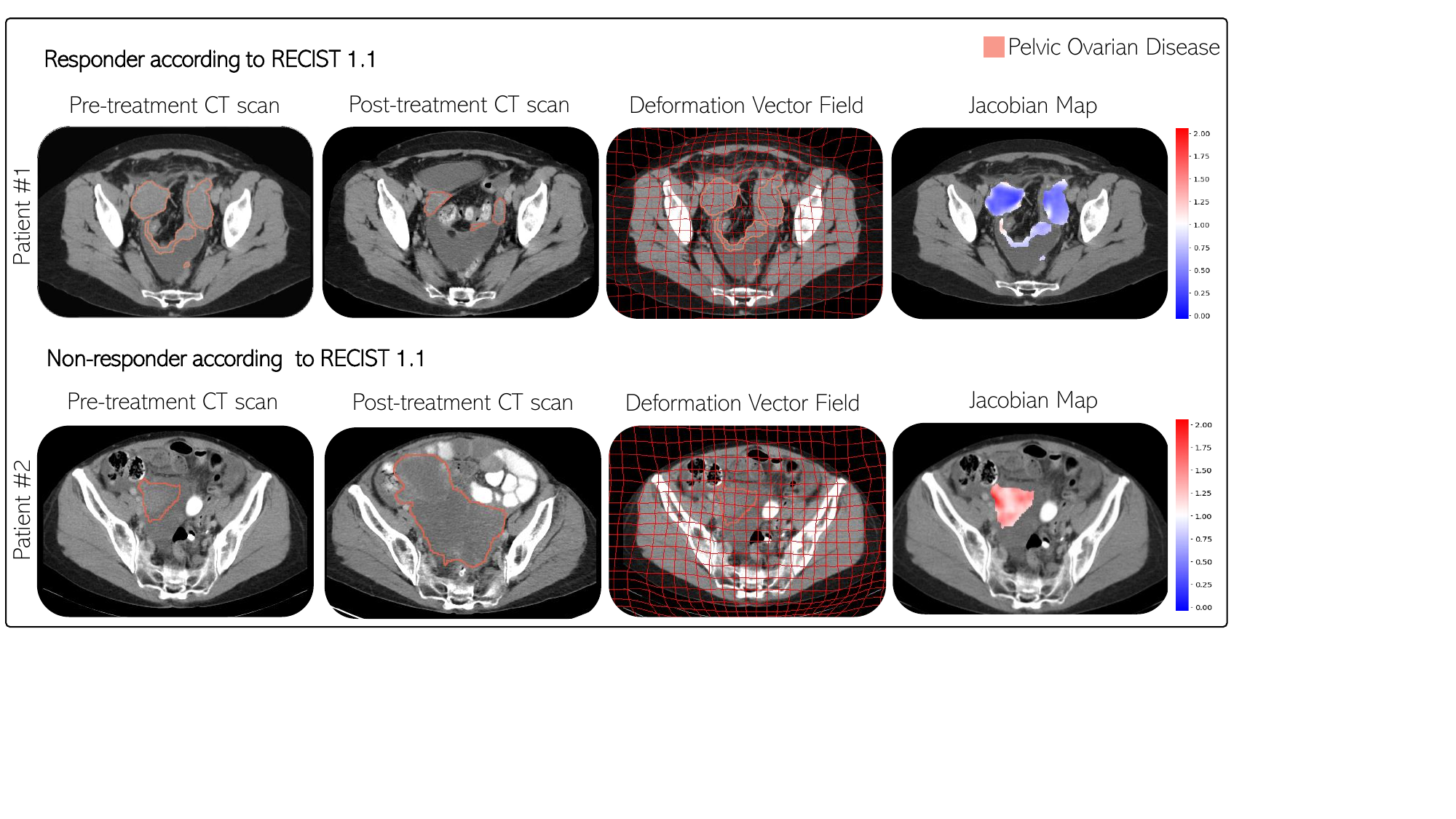}
\caption{Baseline, follow-up, DVF, and Jacobian maps in axial view for HGSOC patients classified as a responder (top row) and a non-responder (bottom row) according to RECIST 1.1. Tumour segmentations are overlaid on top of both the DVFs and the Jacobian maps, showing longitudinal volume changes. Red and blue colours in Jacobian maps represent expansion and shrinkage respectively, whereas white colour shows volume preservation.} \label{jacobs}
\end{figure}

\noindent\textbf{Quantification of tumour deformation.} Table 1 (C) shows the average median Jacobian for all disease sites, and separately for omental and pelvic ovarian disease, as these comprise 90.8\% of the total tumour volume. Results are stratified by macroscopic areas, including hypo-dense (i.e., cystic/necrotic), hyper-dense (i.e., calcified), and intermediate dense (i.e., soft tissue) portions and categorised by patient subgroups based on their treatment responses. Figure~\ref{jacobs} shows the baseline image, the follow-up image, the deformation vector field and Jacobian map in axial view for HGSOC patients classified as a responder (top row) and a non-responder (bottom row) according to RECIST 1.1. In the top row, pelvic ovarian disease shrinkage is visible in the follow-up image. Tumour shrinkage is illustrated by the DVF overlaid on top of the baseline image (red contour). Converging vectors generated a sink point in the middle of the tumour region where median Jacobians were smaller than 1 (dark blue) indicating shrinkage. Quantitatively, the majority of voxels (97.53\%) were shrinking with average Jacobian of 0.46±0.15, indicating overall a 54\% shrinkage in the tumour volume. The bottom row shows a non-responder tumour with its DVF and Jacobian map. In the follow-up image, a large expansion appeared on the left distal side of the pelvic ovarian lesion. DVF overlaid on the baseline image illustrates that vectors are diverging toward the right distal side of tumour (red contour) where median Jacobians were larger than 1 (dark red), indicating expansion. In this case, majority of voxels (98.26\%) were expanding with Jacobian greater than 1. The mean Jacobian was 1.34±0.18, indicating overall a 34\% expansion in the tumour volume. Figure 5 shows examples of tissue-specific sub-segmentations. The whole tumour, hyper-dense and hypo-dense region contours are displayed as light blue, yellow and magenta solid lines, respectively. 

\vspace{-3mm}
\section{Discussion}
\vspace{-1mm}
We benchmarked state-of-the-art medical image registration algorithms, including traditional methods (NiftyReg), deep learning-based methods (VoxelMorph), and foundation models (uniGradICON), for the first time in the context of longitudinal CE-CT registration for HGSOC. Table 1 shows that the proposed foundation model outperforms the traditional and DL-based methods in terms of registration accuracy in both tumour and non-tumour regions and speed. We also show that registration accuracy varies across different regions (disease sites vs. anatomical structures) and among patient subgroups with different responses to chemotherapy. Omental disease poses a more challenging registration task compared to the pelvic ovarian disease, as omental lesions are usually numerous and densely packed within a single organ. 

\begin{table}[H]
\resizebox{\textwidth}{!}{
\begin{tabular}{|c|c|c|c|c|}
\hline
\multicolumn{2}{|c|}{Location/Registration Method} & NiftyReg [14min] & VoxelMorph [8s] & GradICON [8s] \\ \hline
\multicolumn{5}{|c|}{\textbf{(A) DSC±SD (SDlogJ) in Tumour Regions}} \\ \hline
\multirow{2}{*}{POD}                    & Responder                        & 0.575±0.089 (0.061) & 0.645±0.042 (0.052) & 0.796±0.261 (0.025) \\ \cline{2-5}
                                        & Non-responder                    & 0.599±0.071 (0.059) & 0.751±0.212 (0.035) & 0.802±0.145 (0.018) \\ \hline
\multirow{2}{*}{OD}                & Responder                        & 0.501±0.073 (0.036) & 0.602±0.056 (0.061) & 0.701±0.383 (0.071) \\ \cline{2-5}
                                        & Non-responder                    & 0.572±0.076 (0.083) & 0.700±0.021 (0.041) & 0.725±0.178 (0.020) \\ \hline
\multicolumn{5}{|c|}{\textbf{(B) DSC±SD (SDlogJ) in Healthy Anatomical Structures}} \\ \hline
\multicolumn{2}{|c|}{Kidneys}                                              & 0.824±0.048 & 0.815±0.067 & 0.913±0.052 \\ \hline
\multicolumn{2}{|c|}{Liver}                                                & 0.801±0.059 & 0.807±0.031 & 0.937±0.076 \\ \hline
\multicolumn{2}{|c|}{Sacrum}                                               & 0.889±0.051 & 0.845±0.078 & 0.901±0.033 \\ \hline
\multicolumn{2}{|c|}{Vertebra}                                             & 0.811±0.086 & 0.978±0.062 & 0.912±0.220 \\ \hline
\multicolumn{2}{|c|}{Colon}                                                & 0.510±0.052 & 0.525±0.043 & 0.720±0.068 \\ \hline
\multicolumn{2}{|c|}{Stomach}                                              & 0.522±0.076 & 0.560±0.035 & 0.779±0.040 \\ \hline
\multicolumn{2}{|c|}{Spleen}                                               & 0.645±0.083 & 0.712±0.028 & 0.800±0.031 \\ \hline
\multicolumn{2}{|c|}{Lungs}                                                & 0.661±0.084 & 0.724±0.084 & 0.889±0.010 \\ \hline
\multicolumn{2}{|c|}{Pancreas}                                             & 0.609±0.035 & 0.739±0.071 & 0.891±0.091 \\ \hline

\multicolumn{5}{c}{} \\ \hline

\multicolumn{5}{|c|}{\textbf{(C) Median Jacobians stratified by Tumour Sub-Regions}} \\ \hline
\multirow{8}{*}[1em]{\centering POD} & \multirow{4}{*}[0.5em]{\centering Responder} & \multirow{4}{*}[0.5em]{\centering 0.80±0.10} & Cyst/Necrotic & 0.918±0.55 \\ \cline{4-5} 
                     &                             &                            & Calcified     & 1.129±0.36 \\ \cline{4-5} 
                     &                             &                            & Soft Tissue   & 0.804±0.39 \\ \cline{2-5} 
                     & \multirow{4}{*}[0.5em]{\centering Non-responder} & \multirow{4}{*}[0.5em]{\centering 1.05±0.15} & Cyst/Necrotic & 1.028±0.29 \\ \cline{4-5} 
                     &                               &                            & Calcified     & 1.215±0.53 \\ \cline{4-5} 
                     &                               &                            & Soft Tissue   & 1.106±0.31 \\ \hline
\multirow{8}{*}[1em]{\centering OD}  & \multirow{4}{*}[0.5em]{\centering Responder} & \multirow{4}{*}[0.5em]{\centering 0.64±0.14} & Cyst/Necrotic & 0.962±0.46 \\ \cline{4-5} 
                    &                            &                            & Calcified     & 0.606±0.21 \\ \cline{4-5} 
                    &                            &                            & Soft Tissue   & 0.640±0.43 \\ \cline{2-5} 
                    & \multirow{4}{*}[0.5em]{\centering Non-responder} & \multirow{4}{*}[0.5em]{\centering 1.02±0.15} & Cyst/Necrotic & 1.004±0.39 \\ \cline{4-5} 
                    &                               &                            & Calcified     & 1.182±0.41 \\ \cline{4-5} 
                    &                               &                            & Soft Tissue   & 1.116±0.39 \\ \hline
\end{tabular}%
}
\caption{Mean and standard deviation of Dice coefficient and the standard deviation of the logarithm of the Jacobian determinant as DSC±SD(SDlogJ), and running time of registration methods across disease sites (A), anatomical structures (B), and patient subgroups. Initial DSC is 0.376±0.45 for tumour regions and 0.546±0.71 for non-tumour regions. (C) Median Jacobian determinants stratified by disease sites - omental disease (OD) and pelvic ovarian disease (POD) - and their macroscopic areas, including hypo-dense (i.e., cystic/necrotic), hyper-dense (i.e., calcified), and intermediate dense (i.e., soft tissue) portions.}
\label{tab:combined-table}
\end{table}

\begin{figure}[H]
\includegraphics[trim={1 9cm 4.2cm 0cm},width=\textwidth]{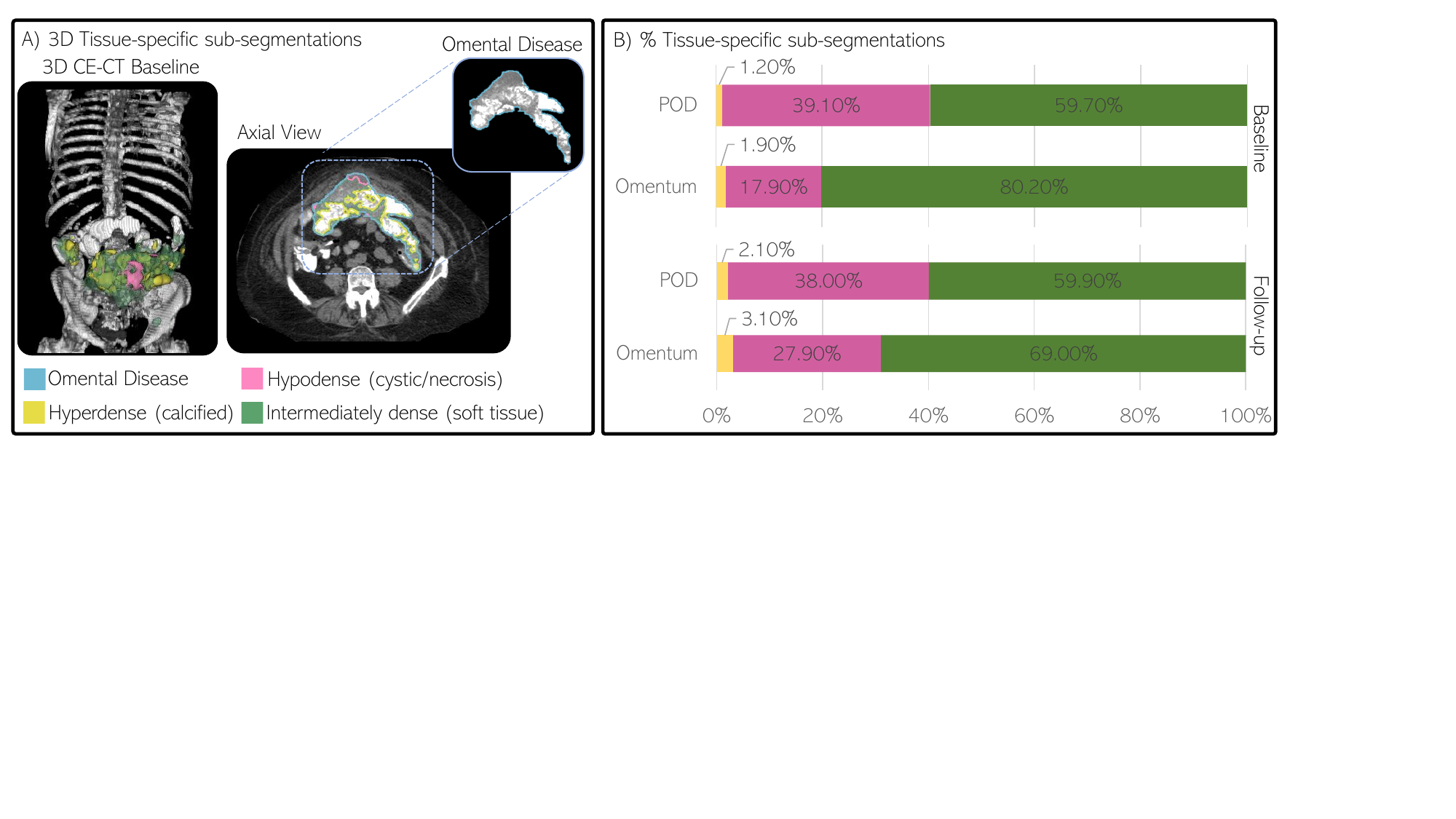}
\caption{(A) 3D and axial CT reconstructions showing segmented ROIs: hyper-dense, hypo-dense and soft tissue components relative to the whole omental tumour. (B) Total percentage of each tissue type for two disease sites at baseline and follow-up for the cohort.} \label{subsegs}
\end{figure}

\vspace{-10mm}
These lesions split or merge over time more frequently than those in the pelvic region, where the number of connected components between pre- and post-treatment scans is generally more stable. Additionally, patients with complete responses exhibit greater tumour changes, posing a more significant registration challenge, as shown by the lowest DSC in Table 1 compared to non-responders. 

Due to the higher accuracy of the proposed foundation model compared to other registration methods, we further evaluated its ability to quantify longitudinal tumour changes induced by NACT. The average median Jacobian for all disease sites was 0.82±0.11 and 1.06±0.15 for responders and non-responders, respectively. These indicated that on average responders had 18\% median volume shrinkage while non-responders had 6\% median volume expansion. For pelvic ovarian disease, the average median Jacobian was 0.80±0.10 for responders and 1.05±0.15 for non-responders, respectively. This indicates that, on average, responders experienced a 20\% median pelvic tumor volume shrinkage, while non-responders experienced a 5\% median pelvic tumor volume expansion. For omental disease, the average median Jacobian was 0.64±0.14 for responders and 1.02±0.15 for non-responders, indicating that, on average, responders had a 36\% median volume shrinkage, while non-responders had a 2\% median volume expansion. Omental disease showed a significantly better response than pelvic ovarian disease. We also quantify, for the first time, localised volumetric deformation in different sub-regions (soft tissue, calcified, and necrotic regions) within tumors. We show that cysts are more prevalent in the POD compared to omental disease, and that calcified regions are typically more prominent in follow-up scans, representing areas of calcified fibrosis or calcified deposits resulting from treatment or disease progression. Table 1 (C) shows that soft tissue shrinks in responders and expands in non-responders, while cyst regions preserve their volume and calcified areas generally increase during NACT.

Upon further evaluation of the performance of registering longitudinal CE-CT images to quantify localised tumour change in HGSOC, our future work includes (a) quantitative evaluation of registration accuracies on longitudinal images with more than one follow-up time point; (b) evaluation of the performance of registration in larger datasets and across a broader range of registration algorithms; and (c) the use of registration-quantified changes and registration-revealed heterogeneity of changes to predict long-term pathological responses to treatment. In conclusion, this work quantifies, for the first time, localised heterogeneous tumour changes across different disease sites in ovarian cancer and demonstrates the ability of registration to characterise spatially heterogeneous effects induced by treatment, which may ultimately be used as more sensitive markers for evaluating treatment response and predicting patient outcomes. 

\begin{credits}
\subsubsection{\ackname} We acknowledge funding and support from Cancer Research UK (A22905) and the Cancer Research UK Cambridge Centre [CTRQQR-2021-100012 and A25177], The Mark Foundation for Cancer Research [RG95043], GE HealthCare, and the CRUK National Cancer Imaging Translational Accelerator (NCITA) [A27066]. Additional support was provided by the National Institute for Health Research (NIHR) Cambridge Biomedical Research Centre [NIHR203312 and BRC‐1215‐20014] and the EPSRC Tier-2 capital grant [EP/T022221/1]. This work was further supported by the backing of the Federal Ministry of Education and Research (BMBF, Grant Nos. 01ZZ2315B and 01KX2021), the Bavarian Cancer Research Center (BZKF, Lighthouse AI and Bioinformatics), and the German Cancer Consortium (DKTK, Joint Imaging Platform). Further support was received through the St. Baldrick's Career Development Grant, NIH R61NS126792, NIH R21TR004265, and NIH R21NS21735. Additionally, we recognize the contribution of the Helmholtz Information and Data Science Academy. The authors would like to acknowledge the support of the Scientific Computing team from the Cancer Research UK Cambridge Institute.

\subsubsection{\discintname} The authors have no competing interests to declare that are relevant to the content of this article. 
\end{credits}

%
%
%
%

\end{document}